\pdfoutput=1
\documentclass[journal]{IEEEtran}

\usepackage{algorithm}
\usepackage{algorithmic}
\usepackage{amsfonts}
\usepackage{cite}
\usepackage[bottom]{footmisc}
\usepackage{color}
\usepackage[final]{trackchanges}
\usepackage{ifpdf}
\ifpdf
  \usepackage[pdftex]{graphicx}
  \DeclareGraphicsExtensions{.pdf,.jpeg,.png}
\else
  \usepackage[dvips]{graphicx}
  \DeclareGraphicsExtensions{.eps,.ps}  
\fi

\renewcommand{\initialsOne}{Angela}
\renewcommand{\initialsTwo}{Barak}
\renewcommand{\initialsThree}{Dan}

\usepackage[cmex10]{amsmath}

\usepackage{array}

\usepackage{url}

\begin{document}
%
\title{A Little More, a Lot Better: Improving Path Quality by a Path Merging Algorithm}
%
%
%

\author{Barak~Raveh,
  Angela~Enosh
  and~Dan~Halperin%
\thanks{B. Raveh is with the  Department of Microbiology and Molecular Genetics, Hadassah
Medical School, The Hebrew University, Jerusalem, Israel}
\thanks{B. Raveh, A. Enosh and D. Halperin are with School of Computer Science, Tel-Aviv
University, Israel e-mail: \{barak,angela,danha\}@post.tau.ac.il}}%

\maketitle

\begin{abstract}
Sampling-based motion planners are an effective means for generating
collision-free motion paths. However, the quality of these motion
paths (with respect to quality measures such as path length,
clearance, smoothness or energy) is often notoriously low,
especially in high-dimensional configuration spaces. We introduce a
simple algorithm for merging an arbitrary number of input motion
paths into a hybrid output path of superior quality, for a broad and
general formulation of path quality. Our approach is based on the
observation that the quality of certain sub-paths within each
solution may be higher than the quality of the entire path. A
dynamic-programming algorithm, which we recently developed for
comparing and clustering multiple motion paths, reduces the running
time of the merging algorithm significantly. We tested our algorithm
in motion-planning problems with up to 12 degrees of freedom. We
show that our algorithm is able to merge a handful of input paths
produced by several different motion planners to produce output
paths of much higher quality.

\end{abstract}

%
\IEEEpeerreviewmaketitle

\section{Introduction}
 \label{sec:Introduction}
\IEEEPARstart{F}{or} several decades, extensive efforts have been
devoted to research of deterministic and probabilistic methods for
finding collision-free paths for moving objects among obstacles,
with applications in diverse domains such as robotics, graphical
animation, surgical planning, computational biology and computer
games~\cite{Choset05, Latombe91, Latombe99, Lavalle06}.
Probabilistic sampling-based methods for motion planning were shown
to be particularly useful when the configuration space of the moving
body has a large number of degrees of freedom~\cite{Choset05,Hsu99,
Kavraki96, Lavalle06, Lavalle00, SongThomasAmato03}. In many
applications, it is also relevant to find a path that is better than
other alternative paths. However, finding an optimal path with
respect to various quality measures is often NP-hard, even in
low-dimensional cases where an arbitrary path can be found
efficiently~\cite{CannyReif87,Asano96,ReifWang98}. Finding
high-quality paths in higher dimensions is even harder.

\subsection{Manuscript Outline}
\label{subsec:outline} In Sections~\ref{sec:RelatedWork}
and~\ref{sec:Contribution} we discuss related work and outline the
contribution of this study. In Section~\ref{sec:Algorithm} we
introduce the basic concept of merging multiple motion paths into a
hybrid-path with improved quality. We outline three variants of the
algorithm, from a na\"{\i}ve implementation, to an asymptotically
faster variant, which makes use of our recently introduced
dynamic-programming edit-distance algorithm for matching pairs of
motion pathways. In Section~\ref{sec:Experiments} we report on
experiments of increasing difficulty, in which we compare the
performance of our approach to exisiting motion planners, for
problems with up to 12 degrees of freedom, where our method is
particularly effective.

\section{Related Work}
\label{sec:RelatedWork}

 We make a distinction between finding
an optimal path with respect to the entire (continuous)
configuration space, as opposed to the restricted formulation, where
we look for the optimal path in a discrete graph of configurations.
The latter is generally much easier\footnote{Although some specific
graph-search formulations are also NP-hard, and call for
pseudo-polynomial-time algorithms or appropriate
heuristics~\cite{Mitchell04}}. In fact, all common deterministic and
sampling-based algorithm aim to capture the topological connectivity
of the free space ($C_{\rm{free}}$) by a discrete graph structure
(the \emph{roadmap} graph), and the final solution is extracted by
searching in this graph using Dijkstra's algorithm~\cite{Dijkstra59}
or one of its variants (such as a maximal bottleneck-clearance path,
retrieved by minute modifications to Dijkstra's algorithm or by
efficient alternatives~\cite{Kaibel06}). Kim \emph{et
al.}~\cite{Kim03} devised an augmented version of Dijkstra's
algorithm for finding the optimal path in a graph, where diverse
optimality criteria can be combined in a flexible manner. Hence, a
major challenge remains to build a representative graph structure
that contains high-quality paths in $C_{\rm{free}}$ to begin with.

 Finding optimal paths for instances of many motion
planning problems is NP-hard~\cite{Mitchell04}, and efficient
analytic solutions were devised only for extremely simple ones, such
as translating objects among polygonal obstacles in the plane using
the shortest path~\cite{Nil69,LeeLin86,Hershberger97optimal} or the
highest-clearance path~\cite{OduYap82}. Already for translating
polyhedra, finding the shortest possible path was shown to be
NP-hard~\cite{CannyReif87,Mitchell04}. Approximation algorithms have
been devised for several NP-hard motion-planning problems; see,
e.g., ~\cite{Asano96,Mitchell04,Papadimitriou85}. As an alternative
to exact solutions, sampling-based algorithms record the
connectivity of $C_{\rm{free}}$ in a discrete graph structure by
connecting random landmarks, and were shown to be probabilistically
complete. Nonetheless, we usually cannot hope to completely cover
the space in reasonable running time using these methods. In
particular, for configuration spaces of high dimension, it might be
difficult to retrieve even a single feasible motion path, not to
mention a high-quality one.

\paragraph{Improving path quality by modifying the sampling algorithm}
Wilmarth \emph{et al.}~\cite{Wilmarth99} improved the local
clearance of sampled configurations by sampling closer to the medial
axis. Nieuwenhuisen \emph{et al.}~\cite{NieuwenhuisenO04} improved
the optimal path length in probabilistic roadmaps by closing cycles
only when they significantly reduce the (graph) path length between
configurations, and Geraerts \emph{et al.}~\cite{Geraerts06IEEERSJ}
combined both approaches. In contrast to the above techniques, the
approach we present below is not tailored for any specific criterion
of path quality, and is designed to allow general formulations of
path quality.

\paragraph{Improving path quality by post-processing} Two
paths are said to be homotopy equivalent if one path can be
 continuously deformed into the other, without introducing any collisions
 along the way. Often the output path of a
 roadmap is homotopy equivalent to another higher-quality path.
 In this case, post-processing procedures ignore the roadmap that
 originally created the path, and focus on small perturbations that improve the
 path within its homotopy class. Path pruning and shortcut heuristics
are common post-processing techniques for creating shorter and
smoother paths, with little chance of switching between homotopy
classes. Geraerts \emph{et al.}~\cite{Geraerts04} locally improved
path clearance using a retraction schemes that resembles the
approach taken by Wilmarth \emph{et al.}~\cite{Wilmarth99}, and more
recently~\cite{Geraerts07}, improved both path length and path
clearance simultaneously (but not other criteria of path quality).
Geraerts \emph{et al.}~\cite{Geraerts07corridors} locally improved
path quality within a corridor (an inflated path) by applying a
force-field to the moving body within that corridor. In this case,
the output path is restricted by construction to the selected
corridor. In the Appendix, we discuss some more related work that
deals with the very formulation of path-quality measures.

\section{Contribution}
\label{sec:Contribution} 
\begin{figure*}[!t]
  \centering
  \includegraphics[width=15cm]{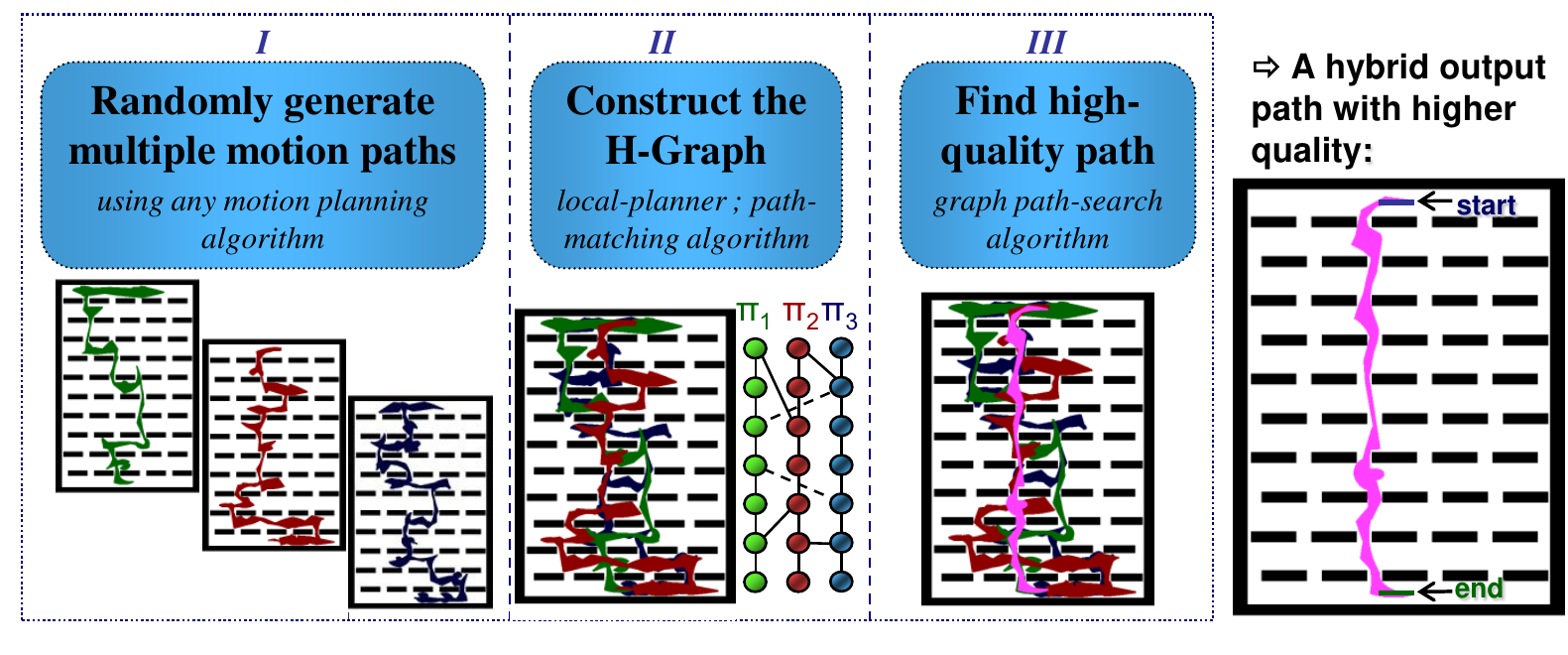}
  \caption{\change[Barak]{older text}{An illustration of using hybridization-graphs to improve
  path quality in the \textit{Grid} scene (extended from~\cite{NieuwenhuisenO04}).
  We visualize the trace of a elongated rectangle that moves with both translation and
  \emph{rotation} in a grid of obstacles}:
  (I) We look for a short path that traverses the grid by first generating
  $l$ paths using any standard-motion planner, either multi-query
    or single-query (PRM in this example). Standard techniques
  often tend to generate lengthy zigzagging motion paths.
  (II) For constructing the H-Graph, we invoke a local-planner
  between certain pairs of configurations. The running time may
  depend on the number of pairs, see Section~\ref{sec:Algorithm}.
  (III) The specific choice of a graph-search algorithm depends on
  the quality measure of interest (output path in magenta). Images
  and PRM paths were generated using the OOPSMP motion-planning package~\cite{PlaBekKav2007:ICRA_OOPSMP}.
  The figure is best viewed in color.
  }
  \label{fig:1}       
\end{figure*}

We observe that sampling-based algorithms
like Probabilistic Roadmaps~\cite{Kavraki96}, Rapidly-exploring
Random Trees~\cite{Lavalle00} and Expansive Space Trees~\cite{Hsu99}
tend to generate different solutions in different runs, depending on
the random decisions made at each run. Output solution paths may
differ by continuous homotopic deformations, but they can also come
from different homotopy classes. See, for example,
Figure~\ref{fig:1}a, where three different paths are shown, produced
by three runs of the PRM (Probabilistic Road-Maps) algorithm.

Planning arbitrary motion-paths is often easier than finding
high-quality paths~\cite{Mitchell04}, and a common practice in
optimization theory is to integrate existing solutions into a new
and improved solution, as is done, for example, in genetic
algorithms~\cite{Holland75}. Following this line of thought, we
observe that even if the entire path has low quality, some shorter
subsegments within the path might be of high quality.

In this study, we describe a simple and efficient post-processing
approach for improving the quality of the motion paths by
hybridizing high-quality sub-paths from initial solutions to the
motion query. The initial solutions can be generated using any
standard single-query or multi-query algorithm for motion planning
(Figure~\ref{fig:1}). We integrate these solution paths within a
graph data-structure, which we call the
\textit{Hybridization-Graph}, or \textit{H-Graph}. We present
several approaches for efficiently merging multiple motion paths
into a single high-quality paths, and show how this simple paradigm
can efficiently produce high-quality paths under various quality
measures, in a general and uniform manner, and without the need for
ad-hoc optimization. This allows for particular flexibility and
improved running time in multi-query settings where the type of
quality criterion may vary between queries, and we avoid recomputing
the entire query each time.

\section{Algorithm}
 \label{sec:Algorithm}

\subsection{Hybridization Graph between Multiple Motion Paths}

Intuitively, sampling-based algorithms (that were not tuned towards
a specific quality measure) are more likely to sample short
high-quality path segments than full length high-quality paths. We
are interested in generating a high-quality output path by
post-processing a set of $l$ collision-free motion paths $\pi_1$,
$\pi_2$, ..., $\pi_l$, which were generated by some arbitrary motion
planner, all sharing the same start and end configuration. The input
paths are assumed to be given as a linear chain of discrete nodes. A
graph $H$ (the \emph{Hybridization-Graph}, or H-Graph) is
initialized from the union of these paths: the vertices $V_H$ are
the disjoint union of intermediate nodes from each path, in addition
to the start and end configurations. Likewise, the edges $E_H$ are
the original edges taken from the $l$ motion paths. After the
initialization of $H$, we use a \textit{local-planner} to connect
certain pairs of configurations that originate from the various
paths\footnote{The local-planner can be considered as a black box
for our purposes, but it typically involves systematic collision
detection of intermediate configurations between a pair of end
configurations}, creating new 'bridges' between or within paths. We
elaborate on the choice of those pairs of configurations below.  The
resulting bridging edges create new paths in $H$ that contain
sub-segments from different input paths. These hybrid paths may have
higher quality than any of the input paths. The pseudo-code for
constructing a general H-Graph is outlined in
Algorithm~\ref{alg:BuildHGraph}. Once $H$ is constructed, we find
the optimal solution with respect to $H$ for any quality measure of
choice, using Dijkstra's algorithm or one of its variants, as
described in the first part of Section \ref{sec:RelatedWork}.

  \begin{algorithm}
    \caption{Building a Hybridization Graph from $l$ input paths}

    {\bf Build-H-Graph(\texttt{PathsList})}
    \begin{algorithmic}
      \STATE \texttt{\textbf{PathsList}:} a set of $l$ input solution paths from
      initial to goal configuration
      \STATE \texttt{\textbf{$G:$}} an output H-Graph
      \STATE
      \STATE initialize-H-Graph(\texttt{PathsList})
      \FORALL{$\pi_1,\pi_2 \in \rm{\texttt{PathsList}}$}
          \STATE \texttt{potentialBridgeEdges} = a list of potential bridging edges between $\pi_1$ and $\pi_2$
          \FORALL{$e \in $\texttt{potentialBridgeEdges}}
        \STATE $\pi_{\rm{local}}$ = localPlanner($e.\rm{\texttt{from}} \rightarrow e.\rm{\texttt{to}}$)
            \IF{valid($\pi_{\rm{local}}$)}
                \STATE $G$.addWeightedEdge($e$)
            \ENDIF
          \ENDFOR
      \ENDFOR
      \RETURN $G$
    \end{algorithmic}
    \label{alg:BuildHGraph}
 \end{algorithm}

\subsection{H-Graph Variants}

A particularly time consuming step in the process of path
hybridization is the invocation of the local-planner, due to
expensive computation such as collision detection queries. We now
discuss several variants of the hybridization algorithm, which aim
to heuristically or asymptotically reduce the number of calls to the
local-planner.

\subsubsection{Exhaustive all-pairs formulation} In this na\"{\i}ve approach
we apply the local planner between all the vertices $V_H$ of $H$
(all originating from the $l$ original paths). If each path consists
of at most $n$ nodes, the local-planner is invoked $O(l^2n^2)$
times, for every pair of configurations and in each pair of paths.
The $n^2$ term is reflected in the list of potential \textit{bridge
edges} in Algorithm~\ref{alg:BuildHGraph}. Since in this variant we
test all pairs of potential bridging edges, the resulting path
quality will be higher or equal compared to the other H-Graph
variants presented below, at the cost of possibly much longer
running time due to exhaustive invocation of the local-planner.

\subsubsection{Neighborhood H-Graph} A simple saving in running time is
achieved by invoking the local-planner only between close-by
configurations in $H$. The neighborhood of a configuration can be
set by a threshold distance $\mathcal{D}$, using an appropriate
distance function on the configuration space. This straightforward
approach can significantly reduce the running time (but also the
resulting path quality), depending on the selected threshold
distance $\mathcal{D}$, and the specifics of the motion planning
problem at hand.

\subsubsection{Edit-Distance H-Graphs}
As mentioned above, in a na\"{\i}ve formulation, the local-planner
is invoked $O(l^2n^2)$ times. We are now interested in reducing the
$O(n^2)$ term by connecting fewer edges. For practical purposes (as
it emerges from our experiments) we can assume that the number of
paths $l$ is small. The Neighborhood H-Graph heuristic, suggested in
the previous section, reduces this number of potential edges but it
is heavily dependant on a neighborhood-size parameter. Another
alternative (which we did not test in the experiments section of
this manuscript) is to invoke the local-planner only for the $\Phi$
nearest neighbours of each node, resulting in an asymptotic
reduction in the number of calls to the local planner from
$O(l^2n^2)$ to $O(ln\Phi)$, but possibly missing many useful
connections. In this section, we take a more structured view and
bound the number of calls to the local-planner for each pair of
paths by $O(n)$, using a recently introduced path matching
algorithm. For completeness of the exposition we briefly describe
the edit-distance algorithm for matching motion
paths~\cite{Enosh08}.

Algorithm \ref{alg:MatchPaths} outlines the details of the dynamic
program for finding an optimal match between discrete motion paths
$p$ and $q$ of lengths $m$ and $n$, respectively.. Let $p_1$, $p_2$,
$\ldots$, $p_m$ and $q_1$, $q_2$, $\ldots$, $q_n$ denote the
configurations along the two paths. We regard each path as a string
with the constituting configurations as letters. A valid matching
between the paths $p$ and $q$ is obtained by aligning
 the two respective strings one on top of the other. Here are examples
 of valid fittings of $p=\{p_1,p_2,p_3\}$ and $q=\{q_1,q_2,q_3,q_4\}$.

\texttt{p1~~p2~~p3~~-  ~~~~~  p1~~- ~~- ~~p2~~p3}

\texttt{q1~~q2~~q3~~q4  ~~~~  q1~~q2~~q3~~- ~~q4}

Formally, in edit-distance string matching we are trying to edit one
string to another string by using \textit{legal operations} of
character replacement and character insertion /
deletion~\cite{Levenshtein66}. We pay a certain price for each of
these operations, and the objective is to find a set of edit
operations where we pay the minimal price. In order to align
matching subpaths, we give a higher penalty for matching
(``replacing'') dissimilar configurations. We also pay a price for
opening a gap in the alignment (``deleting/inserting''), for
mismatching subpaths. In this case, we align a letter against a
space character (denoted above as a dash), but we need to pay a
penalty as well. The edit operations induce a natural pairwise
alignment between the two paths, with gaps in regions of mismatch.

Let $\Delta(p_i,q_j)$ be the price we pay for replacing
configuration $i$ of path $p$ with configuration $j$ of path $q$. As
stated, we pay less for replacing similar configurations, in order
to match similar regions. In addition, we pay a price
GAP$_{\rm{ext}}$ for extending gaps (and optionally, a price
GAP$_{\rm{init}}$ for initiating a new gap in the match\footnote{For
simplicity, in Algorithm \ref{alg:MatchPaths} below we ignore the
GAP$_{\rm{init}}$ price, the price for initiating a new gap. This
term is incorporated by a minor technical modification.}). A large
GAP$_{\rm{ext}}$ penalty prevents gaps and favors the matching of
subsegments even if they are not very similar, whereas a small
GAP$_{\rm{ext}}$ penalty forces a more selective matching. The
GAP$_{\rm{init}}$ penalty favors a small overall number of
non-consecutive gaps (regardless of the length of consecutive gap
stretches).

 The matrix $C$ contains the optimal costs of matching subsegments from
$p$ and $q$. The entry $C_{ij}$ is the suboptimal cost of matching
the first $i$ configurations from $p$ to the first $j$
configurations from $q$. We fill it in by deciding whether to match
the configurations $p_i$ and $q_j$ or to open a gap. We record this
decision in the trace-back matrix $T\!B$. In the last iteration, the
cost of the alignment is in the last entry of $C_{mn}$, and the
trace-back matrix records the set of edit operations that leads to
an optimal matching. Thus, the alignment between motion paths can be
easily recovered from the $T\!B$ matrix.

 \begin{algorithm}
    \caption{Dynamic-Programming Algorithm for Matching Two Paths}

    {\bf MatchPaths($p$,$q$)}
    \begin{algorithmic}
        \STATE \textbf{$C$:} a cost matrix $\in \Re^{m \times n}$
        \STATE ~~~~~~~~\COMMENT{\textit{For $i<0$ or $j<0$, we define $C_{i,j} = \infty$}}
        \STATE \textbf{$T\!B$:} a symbolic trace-back matrix

        \STATE
        \FOR{$i$=0 to $m$}
            \FOR{$j$=0 to $n$}
                \STATE \texttt{Match}$\Leftarrow C_{i-1,j-1} + \Delta(p_{i},q_{j})$
                \STATE \texttt{Up}$\Leftarrow C_{i-1,j} + \rm{GAP}_{\rm{ext}}$
                \STATE \texttt{Left}$\Leftarrow C_{i,j-1} + \rm{GAP}_{\rm{ext}}$
                \STATE $C_{i,j} \Leftarrow \min{(\rm{\texttt{Match}, \texttt{Up}, \texttt{Left}, 0})}$
                \STATE $T\!B_{i,j} \Leftarrow \left\{
                                                         \begin{array}{ll}
                                                           "\nwarrow" & \hbox{for Match} \\
                                                           "\uparrow" & \hbox{for Up} \\
                                                           "\leftarrow" & \hbox{for Left}
                                                         \end{array}
                                                       \right.$
            \ENDFOR
        \ENDFOR
        \RETURN{matrices $C$ and $TB$}
    \end{algorithmic}
    \label{alg:MatchPaths}
 \end{algorithm}

Although the asymptotic running time is clearly $O(mn)$, the
practical running time is negligible compared to calls to the
local-planner between two configuration, where expensive collision
checks are performed.

\emph{Using Path Matching to Speed Up the Construction of H-Graphs.}
We now show how the alignment algorithm is used to reduce the number
of calls to the local-planner during the construction of $H$.
Intuitively, matched subsegments tend to come from close-by regions
of the configuration space (this is reflected in the $\Delta$ cost
function). We observe that gaps between matched regions point to
possible alternative routes. Therefore, we use the path matching
algorithm to bound the number of tested hybridization-edges by
$O(n)$ as follows. In gap regions, we try to connect the ``deleted''
configurations in path $p$ to the two boundary configurations of the
gap in the matched path $q$, using the local-planner. In addition,
we also try to connect matching configurations, in order to obtain
local improvements for the matched region. Since the size of the
alignment is clearly $O(n)$ (where $n$ is the number of nodes in the
longest input path), we try to connect at most $O(n)$ pairs of
configurations for each pair of input paths, in contrast to $O(n^2)$
configurations in All-Pairs H-Graphs.

As in the previous section, a heuristic speed-up in performance can
be achieved by connecting only close-by configuration in the
neighborhood of the match. The combined approach benefits from both
the asymptotic speed-up of $O(n)$ achieved using path matching, and
from the heuristic speed-up achieved by bounding the neighborhood
size. We call this version \textit{Edit-Distance Neighborhood}
H-Graphs.

\section{Experiments}
 \label{sec:Experiments}
In this section, we benchmark the effectiveness of hybridizing
multiple input paths from short runs of sampling-based algorithms,
in a set of 2D, 3D, 6D and 12D configuration spaces. The advantage
of using H-Graphs becomes apparent in the 6D and 12D spaces, where
exhaustive sampling is not feasible.

We implemented the H-Graph algorithm and the path matching algorithm
within the OOPSMP open-source package for motion
planning~\cite{PlaBekKav2007:ICRA_OOPSMP}, and used the package's
internal implementation of PRM, RRT and subdivision local-planner.
The described algorithms can be readily subjected to
parallelization. However, for the performance analysis, we have run
all tests on a single processor of a dual-core 2.67Ghz AMD Athlon
machine. Our example scenes are mostly borrowed from Geraerts
\emph{et al.}~\cite{Geraerts07}.

test

\begin{figure*}[!t]
  \centering
  \includegraphics[height=9.5cm]{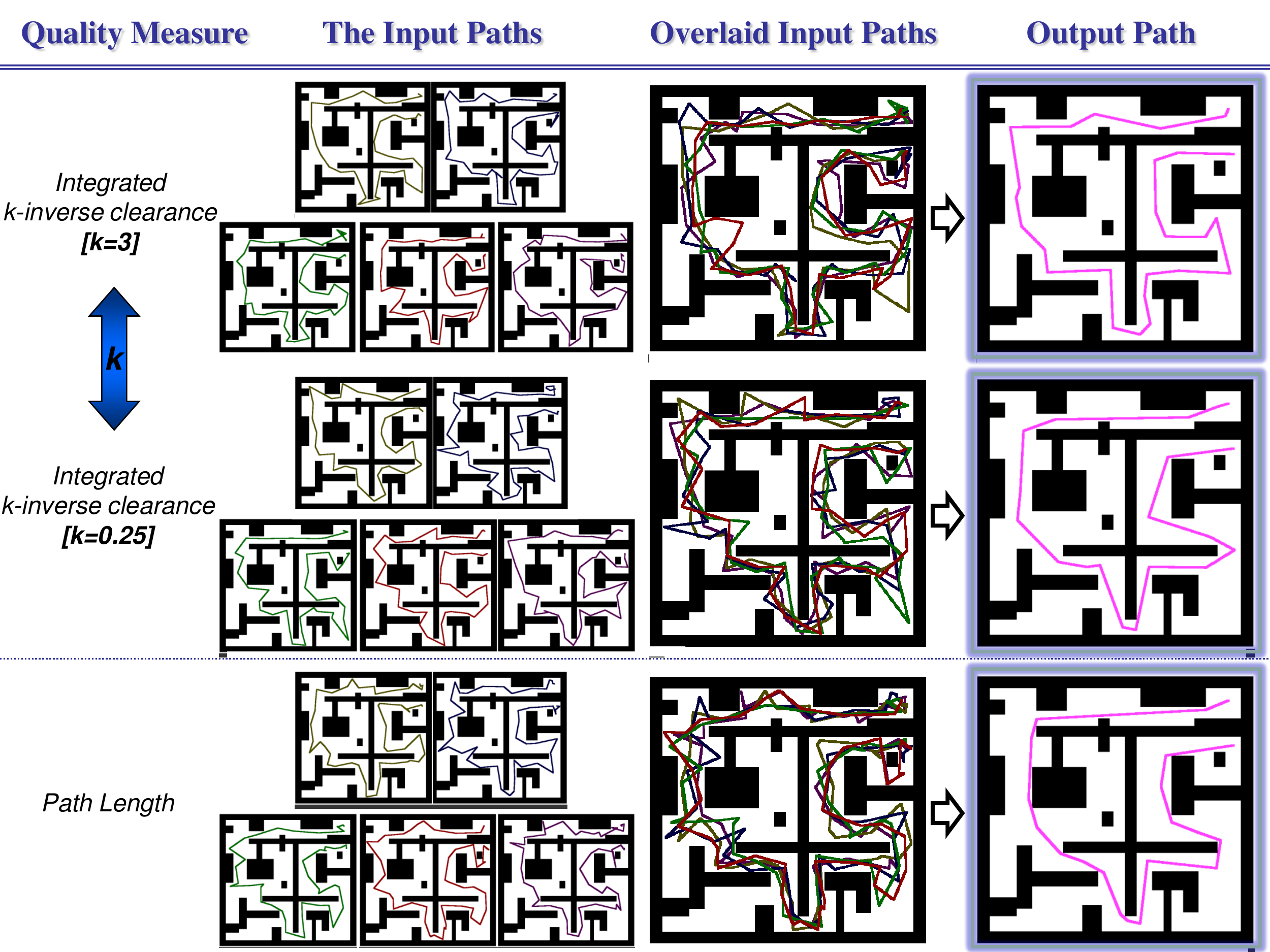}
  \caption{Improvement of various quality measures with H-Graphs in the 2D-Maze scene taken
  from Geraerts \emph{et al.}~\cite{Geraerts07}. Here we integrated five random runs of PRM. Although each input path is already
  the highest-quality path within its roadmap, the output path shows significant improvement
  in path quality.  For instance, in the top rows we try to improve the path length weighted by clearance with different weights
   (See Appendix for more information about the \textit{integrated k-inverse clearance} measure).
  Note that when the clearance is given a high weight (top row), our output hybrid-path is far from the obsacles (right column) although
  all the input paths from which it was made graze the obstacles (left column). As
   expected, when the clearance is given a lower weight (middle row), the \textit{integrated k-inverse clearance} measure
   behaves similarly to the path-length measure (bottom row), but the
   path clearance is higher (right column), illustrating the tradeoff between the two. Input PRM paths (left column) were generated using the OOPSMP motion-planning
   package~\cite{PlaBekKav2007:ICRA_OOPSMP}.
  }
  \label{fig:2}       
\end{figure*}

test2

\subsection{2D Maze: Trading Off Path Length and Path Clearance}
The maze scene is an illustrative toy-example for the flexibility
with which H-Graphs accommodate diverse types of quality measures in
a uniform manner. The scene comprises a small square robot that is
translating in a 2D maze (Figure~\ref{fig:2}) through a corridor
flanked by branching paths that all lead to dead-ends. In
Figure~\ref{fig:2} we see five paths from different PRM runs,
hybridized using various quality measures. In solutions generated by
PRM, the clearance from the maze walls fluctuated considerably
(second column of Figure~\ref{fig:2}). We used the
\textit{Integrated k-Inverse Clearance} (the path length weighted by
the exponentiated inverse clearance $Cl^{-k}$ of configurations ;
see Appendix for further details). Setting $k$ to $3$ gives a fairly
high penalty for regions of low clearance, whereas for
$k=\frac{1}{4}$ path-length also plays an important factor. Indeed,
for $k=3$ we get a high-clearance path. We compare to the
\textit{path-length} measure, where the output path closely
resembles the optimal path. We also experimented with the
average-clearance and the maximal bottleneck clearance (see Appendix
for definition of both quality measures; results not shown). In all
cases, the objective quality measure was improved significantly,
using a very small number of input paths, demonstrating the
flexibility of the path hybridization scheme.

\begin{figure}[!t]
  \centering
  \includegraphics[width=8.5cm]{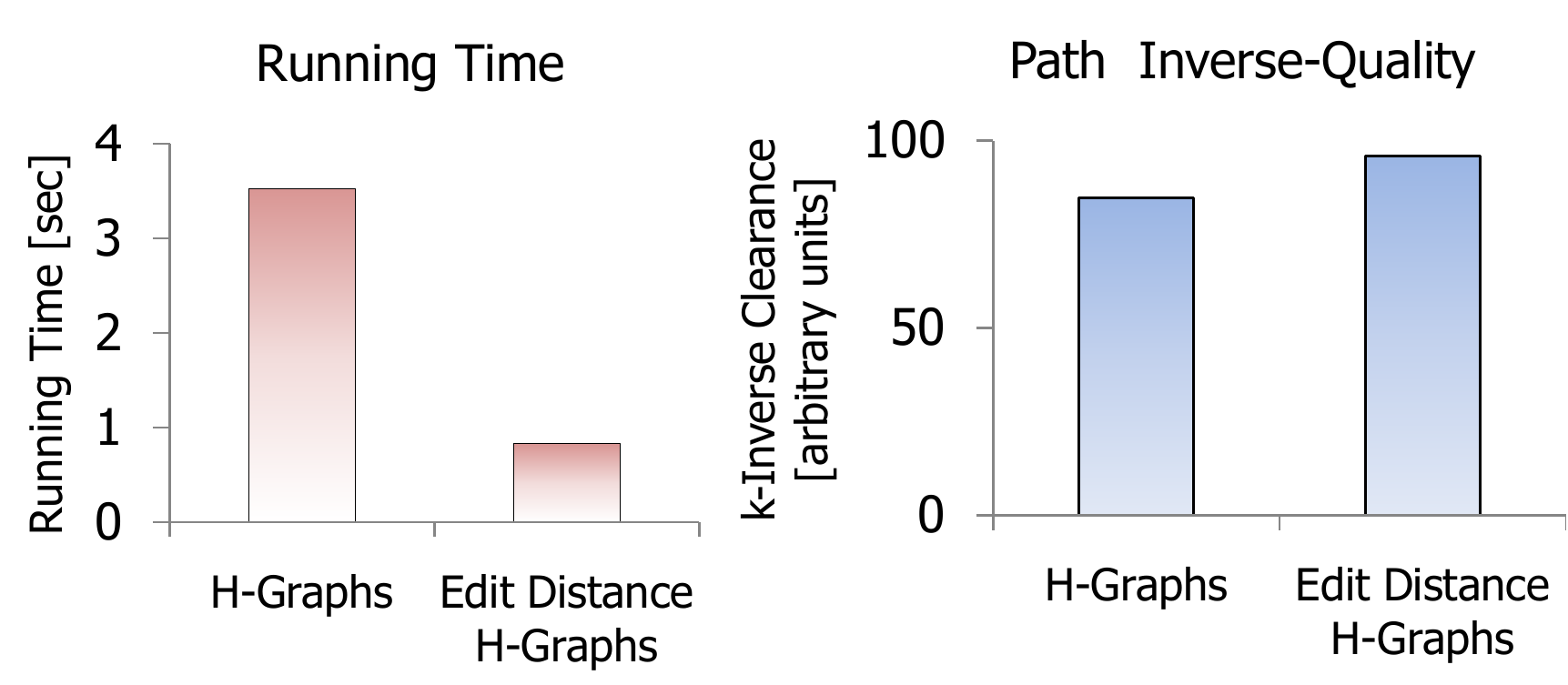}
  \caption{Performance comparison of the \textit{Neighborhood Edit-Distance} vs. the \textit{Neighborhood Edit-Distance}
  variants for hybridizing five input PRM paths, in the 2D maze scene.
  The \textit{Integrated k-Inverse Clearance} is computed with $k=3$ (weighting the path length
  by the clearance of configurations - a lower value means better quality; see Appendix for details on quality measures). }
  \label{fig:maze_bench}       
\end{figure}

\emph{Performance tradeoff with Edit-Distance H-Graphs} As described
in the previous section, the edit-distance path matching algorithm
bounds the number of calls to the local-planner, but the quality of
the path might be compromised because less hybridization-edges are
created. For the 2D-maze environment, we compared the
post-processing time and the clearance quality for hybridizing five
paths of \textit{Neighborhood H-Graphs} and \textit{Neighborhood
Edit-Distance H-Graphs} using the \textit{Integrated k-Inverse
Clearance} measure (path length weighted by clearance) with $k=3$.
Importantly, the running time of the edit-distance variant dropped
by 75\% (Figure~\ref{fig:maze_bench}), while the quality of output
paths was similar between the two variants, although slightly lower
for the Edit-Distance variant.  It is interesting to note that
similar results were obtained for RRT inputs (results not shown),
demonstrating the modular nature of the H-Graph algorithm, and its
independence from the source of input.

\subsection{Elongated Rectangle in a 2D Grid: Rotation and
Translation} In the grid scene, inspired by a scene in Nieuwenhuisen
\emph{et al.}~\cite{NieuwenhuisenO04}, we move an elongated
rectangle with rotation and translation, from the top of the
workspace to the bottom through a grid of rectangles
(Figure~\ref{fig:1}). In each row of obstacles, the elongated
rectangle is forced to squeeze through one of four narrow passages.
In order to make it harder for the robot to cross two consecutive
passages, we distort the grid slightly such that passages at every
second row are shifted with respect to odd rows. Standard
sampling-based algorithms tend to move the robot in a lengthy
zigzagging motion.

We generated three input paths with PRM for the grid scene, and
hybridized them using the \textit{Neighborhood H-Graph} variant, as
illustrated in Figure~\ref{fig:1}. For measuring path length, we
give a relatively large weight to the translational component of the
motion\footnote{We define the translational weight to 1 and the
rotational weight of $0.00005$ in the OOPSMP software . Note the the
units for rotation and translation are incomparable, and in
practice, rotation weights in the order of $10^{-3}-10^{-4}$ are
considered extremely high.}. In the output path of the H-Graph, the
average path length was $1.08 \pm 0.12$ units. In comparison, the
input paths generated with PRM have been well over three times
longer ($3.85 \pm 0.75$ units on average) and the the hybridized
output path shown in Figure~\ref{fig:1} is $0.85$ units long.

\subsection{Comparison of H-Graphs to Long Runs of Various Motion-Planners}
 We showed how H-Graphs can improve the output of a few short runs of PRM. But if
we allow infinite time for each PRM run, then due to its
probabilistic completeness, we would eventually cover the space and
find the approximate shortest path in the roadmap. We now compare
the performance of H-Graphs based on several short runs of PRM, to a
single longer run of PRM. We show that H-Graphs are particularly
effective for high-dimensional configuration spaces, in which
exhaustive runs of PRM and RRT are not feasible.

\begin{figure}[!t]
  \centering
  \includegraphics[width=8cm]{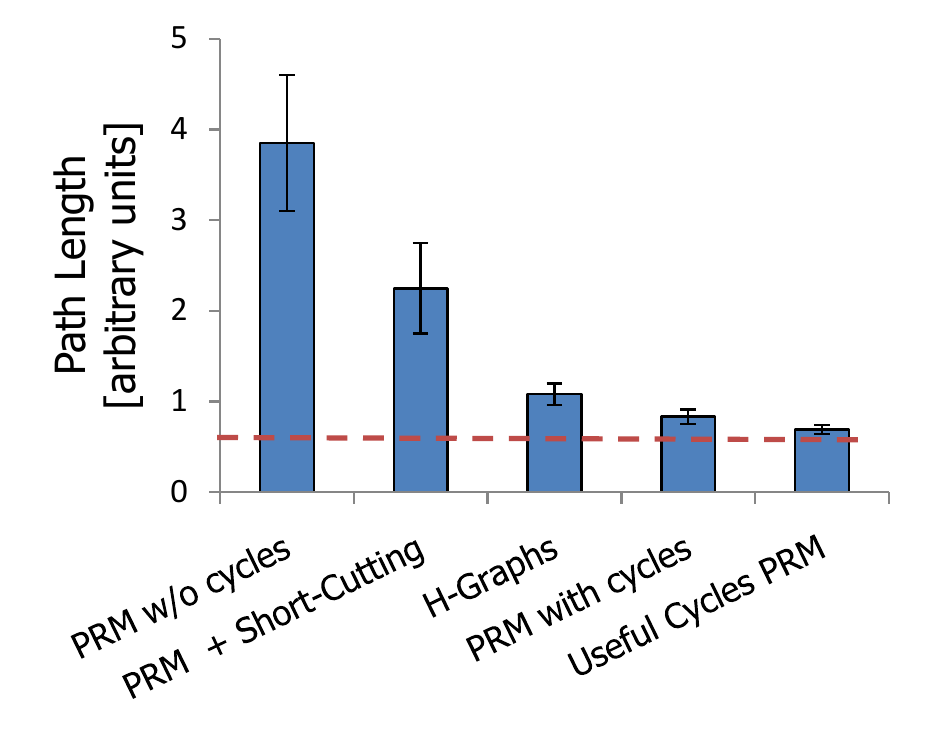}
  \caption{
  Comparison of output path lengths for the grid scene (three degrees of freedom) between different motion planners and the Hybridization-Graphs approach.
  Identical running times ($3.4$ seconds) were allocated for each motion planner. The
  dashed line indicates the approximate optimal path length
  (estimated by extremely long runs of PRM with cycles).
  }
  \label{fig:rod_bench}       
\end{figure}

\emph{PRM without cycles:} In the grid environment described above,
which has three degrees of freedom, we hybridize three PRM input
paths. We allocate $1$ second for preprocessing each PRM, and
together with $0.4$ seconds for constructing the H-Graph on average,
the total running time $t_{\rm{total}}$ is \textbf{$3.4$} seconds.
In comparison, we let PRM run for $t_{\rm{total}}=3.4$ seconds. In
the long PRM run, the average path length is still very long
compared to exhaustively long runs that find near-optimal paths
(Figure~\ref{fig:rod_bench}). Strikingly, even if we more than
double the running time of PRM to $8$ seconds, the shortest path in
the roadmap is still much longer than in H-Graphs ($2.67 \pm 0.37$).

\emph{Short-cutting Heuristics:} It is worth noting that our method
differs significantly from the common short-cutting post-processing
heuristics in which the local planner is applied internally within
the path (described in, e.g., Geraerts \emph{et
al.}~\cite{Geraerts07}). While this heuristics may be useful for
getting rid of certain loops and ``bumps'' in a motion path, it is
generally unsuitable for transforming between different homotopy
classes (except for simple instances of motion planning problems
devoid of narrow passages). Our results confirm this, as PRM paths
in the grid scene that were post-processed with the standard
short-cutting heuristics are over twice longer than hybridized
motion paths (Figure~\ref{fig:rod_bench}).

\emph{PRM with cycles:} A common practice in PRM is to refrain from
adding cycles to the roadmap, since they promote a quadratic
increase in the number of edges, impairing
performance~\cite{NieuwenhuisenO04}. However, the poor quality of
paths in long PRM runs described above, compared to H-Graphs, can be
attributed to the lack of cycles. Indeed, in the same example of the
elongated rectangle that is translating and rotating through the
grid, long runs of PRM with cycles result in slightly shorter paths
compared to H-Graphs (Figure~\ref{fig:rod_bench}). We also
implemented and compared our method to the \textit{useful-cycles}
extension of PRM by Nieuwenhuisen \emph{et
al.}~\cite{NieuwenhuisenO04}, where a cycle is closed if the
connecting edge provides a significant short-cut, by a factor
$\gamma$ (we used $\gamma=3$). In this case, for the same running
time we also get a slightly better path length
(Figure~\ref{fig:rod_bench}). While in this very simple example with
three degrees of freedom, we might be better-off running PRM
\emph{with cycles} for a long time instead of using H-Graphs, we now
demonstrate the advantage of using H-graphs for problems of higher
dimensions.

\subsection{H-Graphs are Effective in 6-dimensional and 12-dimensional configuration spaces}
\begin{figure*}[!t]
  \centering
  \includegraphics[width=18cm]{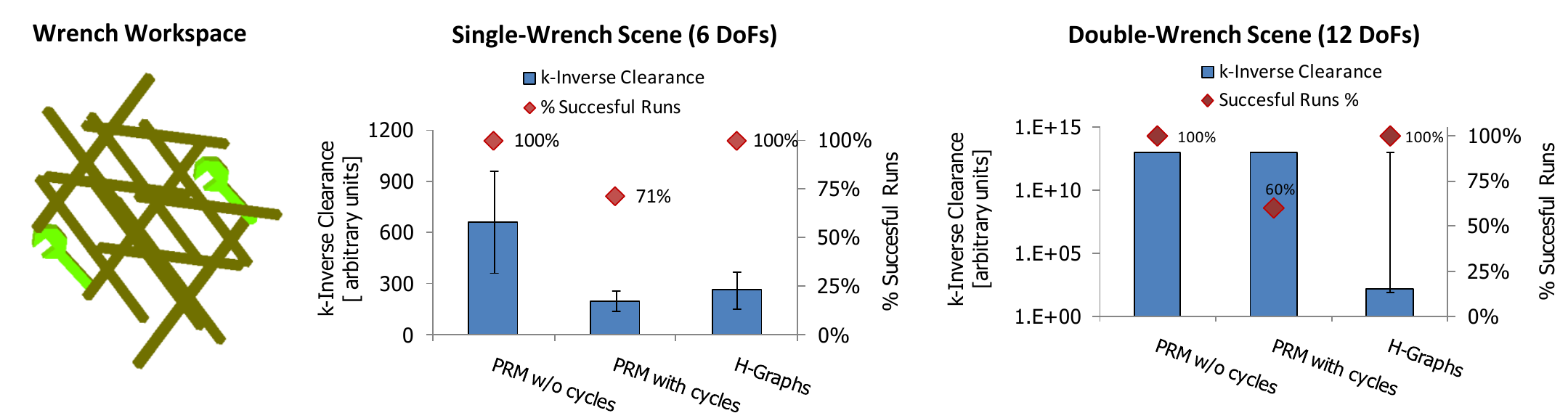}
  \caption{The Single and Double-Wrench Scenes (6 and 12 degrees of freedom ;
  adapted from Geraerts \emph{et al.}~\cite{Geraerts07}).
Results for the Single-Wrench scene are averaged over $20$ runs. We
weighted the path length by clearance using the k-Inverse Clearance
measure with $k=3$ (a lower value means better quality; see Appendix
for details on quality measures). PRM \emph{with cycles} failed to
find a valid solution for 30\%-40\% of the cases in the given time.
For the Double-Wrench scene the median was taken instead of average
in order to avoid the effect of outliers (the error-bars mark the
0.15 and 0.85 percentile quality in each case). Note that the
results for this scene are shown on a logarithmic scale. }
  \label{fig:wrench}       
\end{figure*}

The Single-Wrench scene~\cite{Geraerts07} requires a finely
synchronized rotational and translational motion of a free-flying
wrench moving through 13 axis-aligned beams (see the left hand side
of Figure~\ref{fig:wrench}). The Double-Wrench scene extends this
problem to two wrenches moving simultaneously. In this six and
twelve dimensional problems, the advantage of using H-Graphs becomes
evident. The inverse clearance of the wrenches from the beams was
optimized by hybridizing three paths generated by PRM \emph{without
cycles} (using the \textit{Neighborhood H-Graphs} variant and the
\textit{Integrated k-Inverse Clearance} measure with $k=3$, as
defined in the Appendix), where each run was allocated a total of
$35$ seconds for the Single-Wrench and $900$ seconds for the
Double-Wrench scene..

For both the Single-Wrench and Double-Wrench scenes, the
inverse-clearance measure was improved dramatically with respect to
PRM \emph{without cycles} (Figure~\ref{fig:wrench}). More
importantly, PRM \emph{with cycles} has become prohibitively slow in
30\%-40\% of the cases, failing to find any solution path in the
allocated time frame (Figure~\ref{fig:wrench}). In contrast, our
H-Graphs approach resulted in high-clearance motion paths in all
runs. And while for a single wrench, PRM \emph{with cycles} outputs
paths with marginally better quality (when it finds a solution path
to begin with), in the Double-Wrench scene our hybridization scheme
dramatically outperforms PRM with or without cycles, with respect to
both path quality and the percent of successful runs
(Figure~\ref{fig:wrench}). Our typical output paths allowed a good
safety distance of 10\%-20\% of the wrench width at any point
through the motion, compared to zero clearance for PRM, the latter
outputting non-realistic motion paths for practical purposes.

\subsection{Further Applications of H-Graphs} In this study we
introduced and thoroughly benchmarked a path hybridization scheme in
various 2D and 3D workspaces with up to 12 degrees of freedom. In a
recent study~\cite{Enosh08} we describe a biological example
involving the coordinated motion of a protein with 104 backbone
degrees of freedom, where we applied a preliminary version of the
path-hybridization algorithm to multiple runs of RRT. We note that
in Enosh \emph{et al.}~\cite{Enosh08}, the emphasis was put on the
biological problem at hand, and not on the computational
methodology. In that work, we were able to obtain a low energy
motion path between the initial and goal states. This example is
particularly encouraging as it shows the applicability of H-Graphs
to complex problems with many degrees of freedom, where it is very
hard to optimize the quality of motion paths. In a recent student
workshop in motion planning, students used path hybridization to
significantly improve motion paths of a non-holonomic moving body,
originally generated with C-PRM~\cite{Song01}, extending H-Graphs to
motion planning under kino-dynamic constraints. See movie and more
details in
\url{http://acg.cs.tau.ac.il/courses/workshop/spring-2009/final-projects/non-holonomic-motion-planner-project}.

\section{Conclusions}
We have reported here on a simple algorithm for hybridizing a set of
input paths into an improved output solution. We treat different
quality measures in a uniform manner, and allow modular usage of any
standard motion-planning algorithm. We show experimental results on
2D, 3D, 6D \add[Barak]{and 12D} configuration spaces, indicating
that this new method is particularly useful for such
high-dimensional problems, and offers uniform treatment for various
optimality criteria and motion-planning algorithms.

\appendices
\section{Formulation of Path Quality Measures}
\label{subsec:AppendixQualityMeasures}
 The quality of a motion path can be formulated by human intuition about what a good or convenient path is.
It is natural to choose a short path that keeps a certain distance
from the obstacles. We may also bound the curvature of the path, or
add dynamic time-derivative constraints regarding the velocity of
the moving object. In molecular biology, we often require a
low-energy path. In this paper we mainly describe experiments that
improve the length or clearance of motion paths, and combinations
thereof. However, we note that our algorithm is readily applicable
to other measures of path quality as well.

\subsubsection{Short paths} In a shortest path we minimize the length
$L(\pi)$ of the path $\pi$. For translating rigid bodies, measuring
distance is quite straightforward, but the addition of rotation
complicates matters. Choosing a proper distance metric using either
configuration-space parameters or workspace geometry is discussed
for example in~\cite{Amato98,Kuffner04}. In molecular biology,
distances are often defined over the workspace, using the root mean
square deviation (RMSD) of atom centers between configurations.
Following a standard practice, we use the weighted Euclidean $L_2$
norm over the configuration-space parameters. For a translating
object, this is simply the workspace distance between a fixed
reference point of the object in its different locations. For bodies
that move with rotation, we consider the distance between quaternion
parameters, measured using a bi-invariant distance metric in the
$SO(3)$ topology~\cite{ParkRavani98}
(see~\cite{Kuffner04,YershovaLavalleMitchell2008} for further notes
on distance metrics in $SO(3)$).

\subsubsection{Path clearance} In order to account for safety
distance from workspace obstacles, we may maximize the clearance at
the narrowest passage point along the path, so-called the
\emph{bottleneck-clearance} $BCl(\pi)$. In a graph, the path with
the maximal bottleneck edge is retrieved by a minor adaptation of
Dijkstra's algorithm (or faster alternatives such
as~\cite{Kaibel06}. If we are interested in an estimate of the
behavior over the entire path, we can maximize the \emph{average
clearance} $ACl(\pi)$ along the entire path. Another option is to
locally maximize the clearance by walking along the
\textit{medial-axis} of the free space~\cite{OduYap82}.

\subsubsection{Tradeoff between length and clearance}
High-clearance is unfortunately contradictory to short paths, since
the shortest path often grazes the obstacles~\cite{Lavalle06}.
Therefore, we may relax the requirement for maximal clearance, and
instead seek the shortest path that obeys a certain safety distance
$\verb"C"$ from the obstacles~\cite{Wein07}. \change[Barak]{old
stuff}{Another appealing option is to use the \textit{Weighted
Length} or \textit{Integrated k-Inverse Clearance} measure with
parameter $k$, which also implicitly bounds the path
curvature~\cite{Wein08}. Small clearance is assigned a high penalty
by exponentiating its inverse by some coefficient $k$. The
exponentiated inverse clearance $Cl^{-k}$ is integrated (or summed
for discrete paths), and shorter paths with high average clearance
are favored. } The coefficient $k$ sets the tradeoff between path
length and path clearance. Clearly, if $k=0$ we get the length of
the path. Interestingly, if $k \rightarrow \infty$, the optimal path
is the one with maximal bottleneck-clearance, since $Cl^{-k}$ is
then dominated by configurations with small clearance.

\section*{Acknowledgments}
The authors would like to thank Roland Geraerts and Mark Overmars
for providing us with motion planning scenes ; Lydia Kavraki, Mark
Moll and Erion Plaku to the open-source package OOPSMP for motion
planning~\cite{PlaBekKav2007:ICRA_OOPSMP}. This work has been
supported in part by the Israel Science Foundation (grant no.
236/06), by the German-Israeli Foundation (grant no. 969/07), and
the Hermann Minkowski--Minerva Center for Geometry at Tel Aviv
University.

\ifCLASSOPTIONcaptionsoff
  \newpage
\fi



\bibliographystyle{IEEEtran}
\bibliography{IEEEabrv,references}
%

\end{document}